\documentclass[conference]{IEEEtran}

\renewcommand{\baselinestretch}{0.96}

\usepackage{amsmath}
\usepackage{setspace}		%
\usepackage[pdftex]{graphicx}
\usepackage{booktabs}
\usepackage{multirow}
\usepackage{rotating}
\usepackage{tablefootnote}
\usepackage{ragged2e}

\newcommand{\comment}[1]{}
\usepackage{graphicx}

\let\OLDthebibliography\thebibliography
\renewcommand\thebibliography[1]{
  \OLDthebibliography{#1}
  \setlength{\parskip}{0pt}
  \setlength{\itemsep}{0pt plus 0.3ex}
}

\ifCLASSINFOpdf
\else
\fi
\begin{document}
\title{FPGA-Based CNN Inference Accelerator Synthesized from Multi-Threaded C Software\vspace{-0.1in}}

\author{Jin Hee Kim, Brett Grady, Ruolong Lian, John Brothers$^\dag$, Jason H.~Anderson\\
Dept.~of Electrical and Computer Engineering, University of Toronto, Toronto, ON, Canada\\
$^\dag$Samsung Semiconductor Inc., San Jose, CA, USA\\
Email: \{kimjin14, bgrady, janders\}@ece.utoronto.ca
}

\maketitle

\begin{abstract}
A deep-learning inference accelerator is synthesized from a \texttt{C}-language
software program parallelized with Pthreads.  The software implementation
uses the well-known producer/consumer model with parallel
threads interconnected by FIFO queues.  The LegUp high-level synthesis (HLS)~\cite{canis:legup2} tool
synthesizes threads into parallel FPGA hardware, translating software parallelism into spatial parallelism.  A complete system is generated where 
convolution, pooling and padding are
realized in the synthesized accelerator, with remaining tasks executing
on an embedded ARM processor.  The accelerator incorporates reduced precision, and
a novel approach for zero-weight-skipping in convolution.
On a mid-sized Intel Arria 10 SoC FPGA, peak performance on VGG-16 is 138 effective GOPS. 
\end{abstract}
\IEEEpeerreviewmaketitle

\section{Introduction}

State-of-the-art accuracy results in image recognition, language translation, image-caption generation, and many other tasks are being achieved with deep convolutional neural networks (CNNs)~(e.g.~\cite{DBLP:conf/nips/KrizhevskySH12,DBLP:journals/corr/SimonyanZ14a,DBLP:conf/cvpr/SzegedyLJSRAEVR15}). 
CNN training is very compute intensive, requiring hours, days or weeks of time using state-of-the-art graphics processing units (GPUs). 
Applying a trained CNN to a recognition task, \textit{inference}, can involve billions of operations. Hardware acceleration is particularly desirable for inference, as training is typically done once offline, whereas inference with a trained network is applied repeatedly.  Moreover, there is increased emphasis on performing CNN inference in an embedded-computing context (e.g.~mobile handsets, self-driving cars), where low-power and low latency are important metrics.  In this paper, we focus on acceleration of CNN inference.

CNN inference has been accelerated with GPUs, custom ASICs, and recently, field-programmable gate arrays (FPGAs).  At present, it is unclear which IC media will ultimately prevail as best for CNN inference acceleration.  However, the speed at which CNN research is evolving, as well as recent research on low-precision CNNs~\cite{DBLP:conf/nips/HubaraCSEB16} bodes well for FPGA technology.  With FPGAs, the accelerator architecture and its datapath widths can be precisely tailored to the target CNN, whereas an ASIC or GPU are necessarily over-engineered with fixed-sized datapaths to handle a broad set of precisions.  Moreover, the reconfigurability of FPGAs permits an accelerator design to be adapted to incorporate new research findings as they arise, for example, the ability to achieve high recognition accuracy with 2-bit precision~\cite{TernaryRes16}.  Lastly, high-level synthesis (HLS) is a relatively mature design methodology for FPGAs~\cite{CongHLS}, permitting a software specification of the accelerator to be synthesized into hardware.  HLS lowers NRE costs by allowing design and debugging to proceed at a higher level of abstraction vs.~manual RTL design.      

We apply HLS and use an FPGA to realize a CNN inference accelerator.  The accelerator is described in  \texttt{C} and synthesized with the LegUp HLS framework~\cite{canis:legup2}.  A unique aspect of LegUp is its ability to synthesize software parallelized with the Pthreads standard into parallel hardware~\cite{ChoiPthreads16}.  We leverage the Pthreads synthesis to exploit spatial parallelism on the FPGA. Specifically, we specify the accelerator in software using the producer/consumer parallelization idiom, well known to software engineers. 20 parallel software threads are synthesized by LegUp HLS into streaming hardware comprising compute kernels interconnected by FIFO queues.

The inference accelerator performs key compute-intensive operations: convolution, subsampling (pooling), and padding. Software executing on an embedded on-die ARM processor performs remaining operations to provide a complete end-to-end embedded solution.  The accelerator architecture incorporates novel features for tiling, data-reuse, and zero-weight-skipping, as many CNNs can be pruned without significant loss of accuracy \cite{DBLP:journals/corr/HanMD15}.  The accelerator's computations are realized in reduced precision, specifically 8-bit magnitude + sign format. 
We demonstrate our accelerator on the VGG-16 CNN for image recognition (ImageNet database).  Our contributions are as follows:
\begin{itemize}
    \vspace{-0.06in}
    \item An FPGA-based CNN accelerator synthesized from multi-threaded (Pthreads) \texttt{C} software.  The  software behavior closely resembles the synthesized  hardware, easing design and debugging by allowing it to proceed in software.  
    \vspace{-0.06in}
    \item Generation and exploration of accelerator architectural variants via software/constraint changes alone.
    \vspace{-0.06in}
    \item Analysis of the efficiency of the HLS implementation, in terms of cycles spent, compared to the theoretical minimum for the architecture.
    \vspace{-0.06in}
    \item A novel architecture for zero-skipping; use of reduced-precision arithmetic.
    \vspace{-0.06in}
    \item A complete end-to-end solution for CNN inference, integrated with Caffe for network training.
    \vspace{-0.06in}
    \item 138 GOPS peak effective performance implementing the VGG-16 CNN for image recognition on a mid-sized Arria 10 SoC SX660 20$n$m FPGA.
\end{itemize}

\section{Background}

\subsection{LegUp High-Level Synthesis}

The Pthreads synthesis flow of LegUp HLS is used to synthesize parallel software threads into parallel hardware.  The multi-threaded software is written using the producer/consumer paradigm, where threads represent computational kernels and communicate with one another through FIFO queues~\cite{ChoiPthreads16}. Producer threads deposit computed partial results into output queues, which are then retrieved by concurrently running consumer threads, which perform further processing.  Note that a given thread can be \textit{both} a producer \textit{and} a consumer, receiving inputs from FIFO queues, computing on those inputs, and depositing results to output FIFO queues. The support in LegUp HLS for hardware synthesis of the producer/consumer parallel model is well-aligned with the computational and communication requirements of deep CNN inference.

FIFO queues that interconnect kernels are realized with a LegUp HLS-provided \texttt{LEGUP\_PTHREAD\_FIFO} structure and API, and can be created with user-provided lengths and bitwidths.  
To illustrate the coding style (used heavily throughout our  implementation), the example below shows a function with one input queue, \texttt{inQ}, and one output queue, \texttt{outQ}.  The function body contains an infinite \texttt{while} loop that reads an input, \texttt{inputData} from \texttt{inQ}, performs computation to produce output \texttt{outData}, and deposits into \texttt{outQ}.  \texttt{pthread\_fifo\_read} and \texttt{pthread\_fifo\_write} are API functions to read-from and write-to queues, respectively.   The \texttt{while} loop is pipelined in hardware by LegUp to realize a streaming kernel that accepts new input each clock cycle.

\footnotesize
\begin{verbatim}
void prodCons(LEGUP_PTHREAD_FIFO *inQ, 
    LEGUP_PTHREAD_FIFO *outQ) { 
    ...
    while (1) {
       inputData = pthread_fifo_read(inQ);
       outputData = compute(inputData);
       pthread_fifo_write(outQ, outputData);
    }
}
\end{verbatim}
\normalsize
\vspace{-0.1in}

\subsection{VGG-16 CNN}

The VGG-16 CNN~\cite{DBLP:journals/corr/SimonyanZ14a} %
is used as the test vehicle for our accelerator.
The input to the CNN is a 224$\times$224 RGB image drawn from the 1000-category ImageNet database.  The image is first passed through 13 convolution layers 
interspersed with occasional max-pooling layers,
ending with 3 fully connected layers.
All convolutional filters are $3\times3$ pixels in dimension.  
Before each convolutional layer, the input feature maps are padded with \texttt{0}s around the perimeter.
Max-pooling is done for $2\times2$ regions with a stride of 2. %
ReLU activation is applied in all cases ($y = max(0, x)$, where $x$ is the neuron output).  The VGG-16 network has over 130M parameters and the reader is referred to~\cite{DBLP:journals/corr/SimonyanZ14a} for complete details.

\section{Architecture}

\begin{figure}
  \centering
    \includegraphics[width=0.7\linewidth]{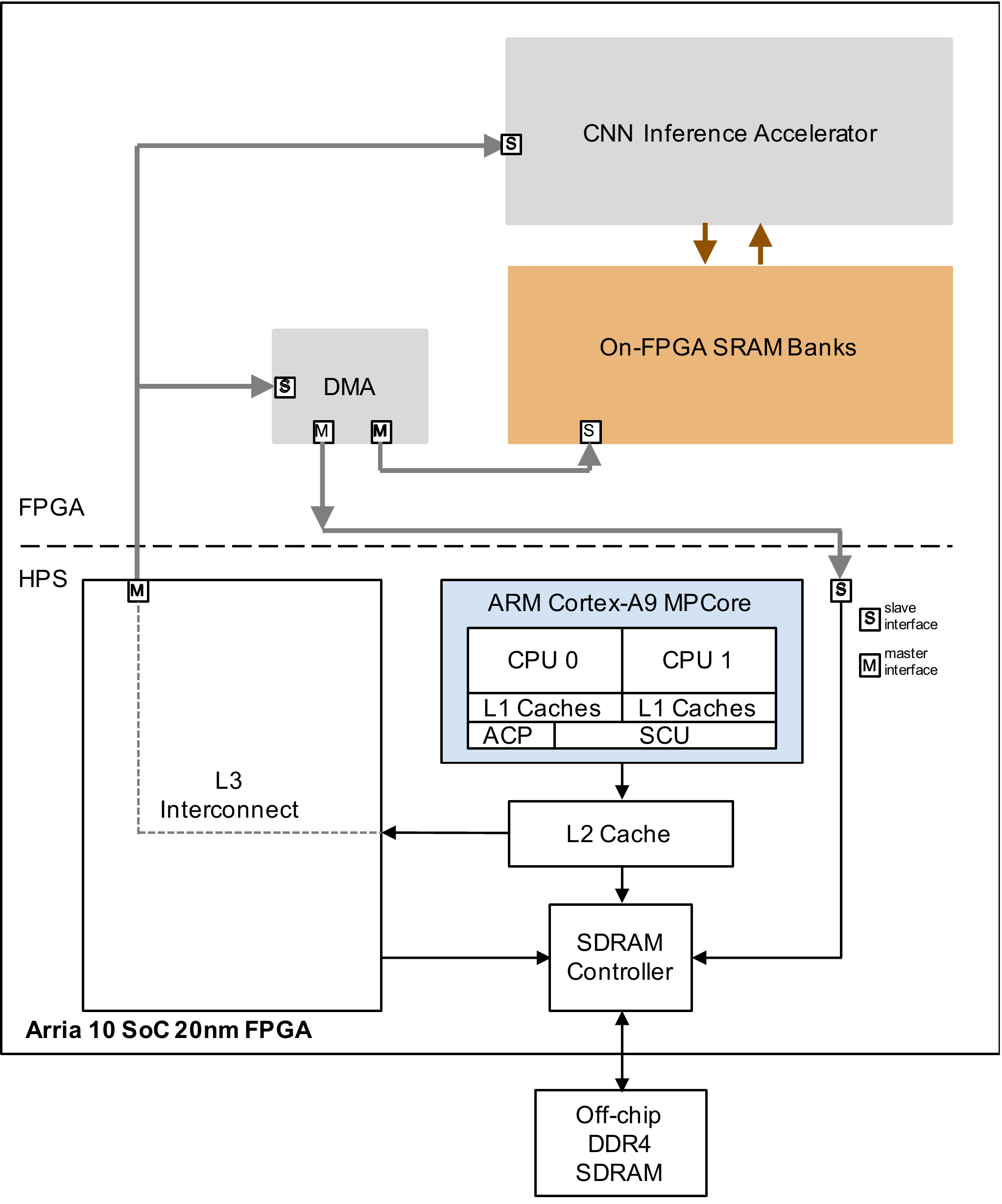}
    \caption{System architecture.}
    \label{fig:system}
    \vspace{-0.1in}
\end{figure}

Fig.~\ref{fig:system} depicts the system-on-chip (SoC) architecture, consisting of a Cortex A9 hard processor system (HPS), accelerator, DMA controller, and SRAM banks within the Arria 10 FPGA fabric. The on-FPGA banks are backed by off-chip DDR4 RAM. The components are connected to one another using Avalon, Intel's on-chip memory mapped bus interface (discussed below). Bus masters are designated with M in the figure; slaves are designated with S. The processor issues instructions to the DMA and accelerator by writing to the memory mapped address (connected through the L3 interconnect). DMA transfers between the off-chip DRAM and FPGA are realized by a direct connection from the DMA unit to the SDRAM controller. 

\subsection{Accelerator Architecture}\label{sec:arch}

\begin{figure}
  \centering
    \includegraphics[width=0.9\linewidth]{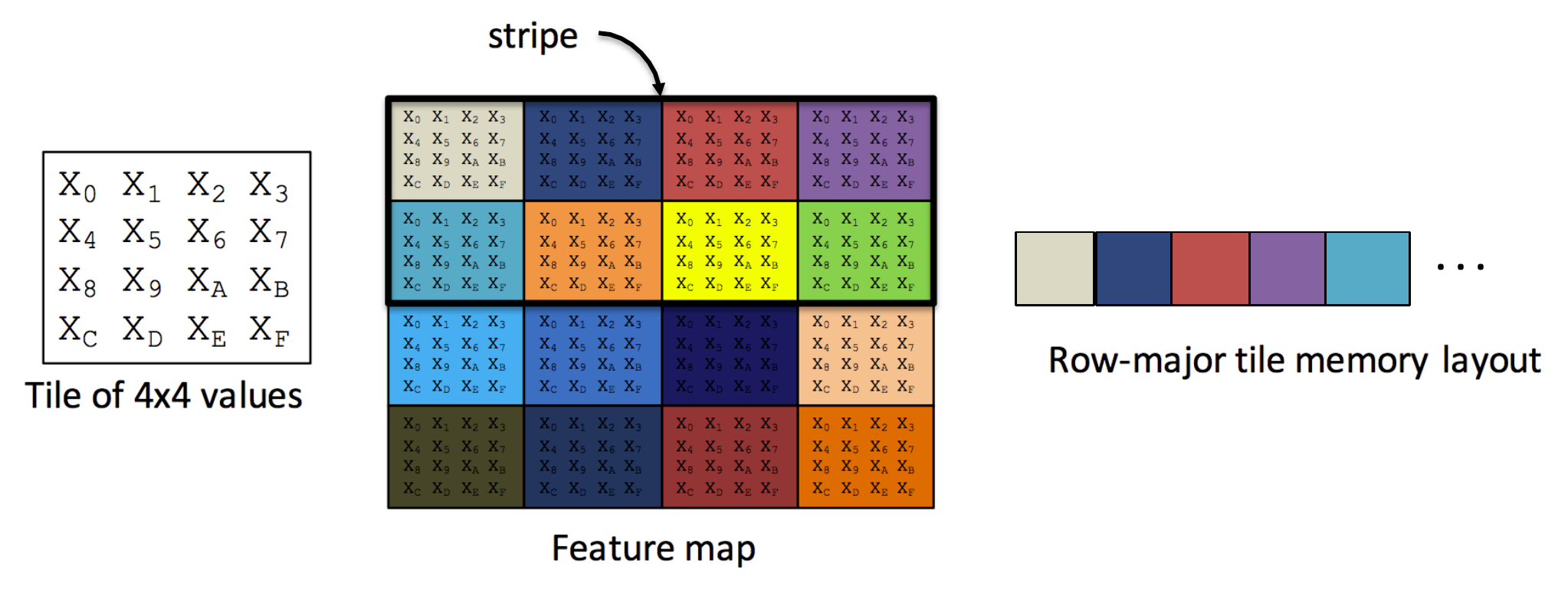}
    \caption{Tile concept, feature map, stripe, data layout.}
    \label{fig:tile}
     \vspace{-0.1in}
\end{figure}

We first %
introduce the data representation, as it is necessary to understand the zero-skipping approach.  Feature maps are organized into tiles of $4 \times 4$ values, as shown on the left side of Fig.~\ref{fig:tile}. The center of the figure shows a $16 \times 16$ feature map, comprising $4 \times 4$ tiles.  These tiles are stored in memory in row-major order, depicted on the right, where colors of tiles correspond to those in the center image.  Fig.~\ref{fig:tile} also introduces the notion of a \textit{stripe}, which is a region of tiles spanning the entire width of a feature map. Striping is used to subdivide large convolutional layers into smaller ones that can be accommodated in on-chip memory.

\begin{figure}
  \centering
    \includegraphics[width=0.9\linewidth]{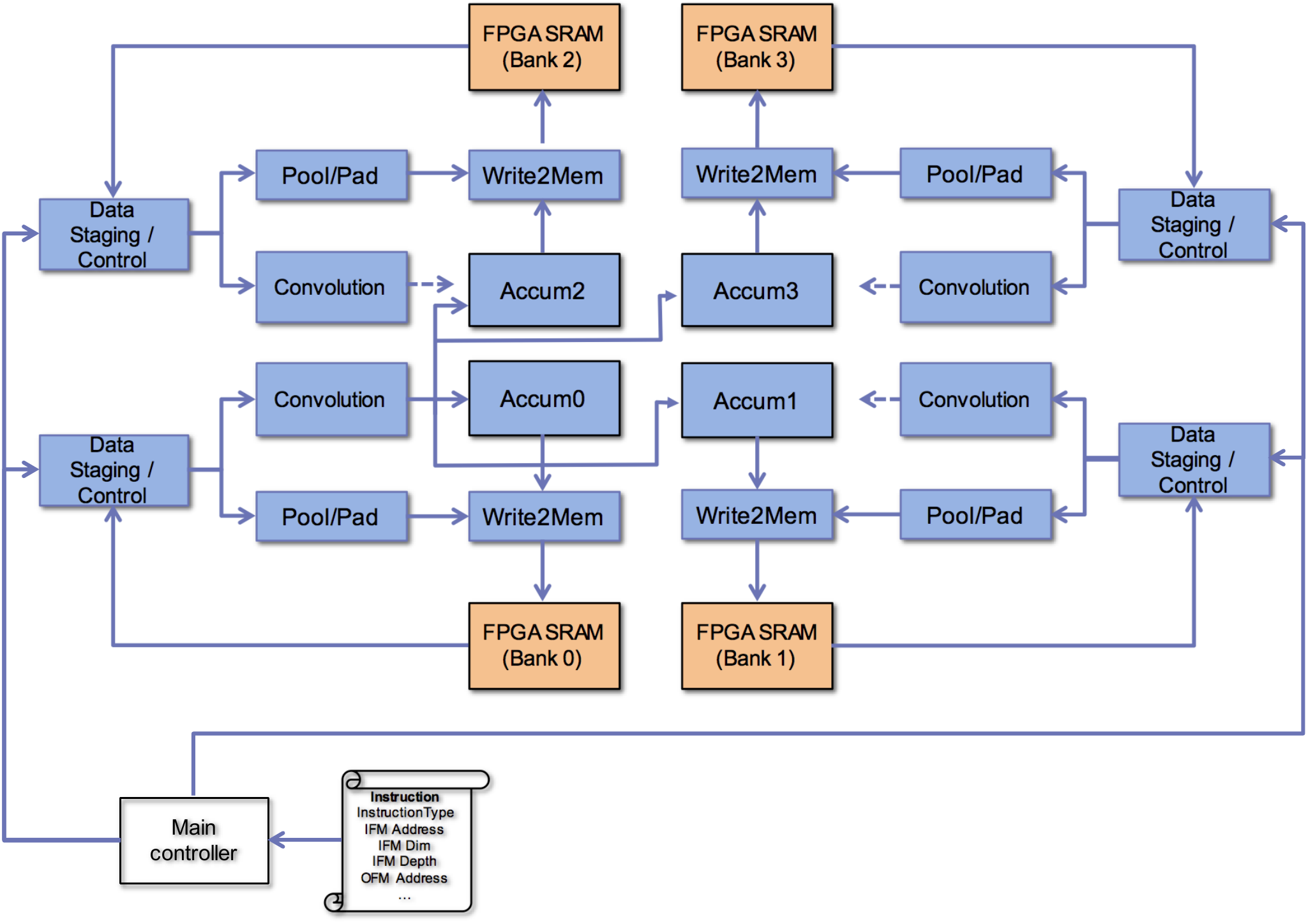}
    \caption{Accelerator block diagram (each blue module is synthesized from  a software thread to hardware).}
    \label{fig:block_dia}
     \vspace{-0.1in}
\end{figure}

A block diagram of the accelerator is shown in Fig.~\ref{fig:block_dia}.  Four banks of on-FPGA SRAM are shown in orange.  An entire tile of data (16 values) can be read from an SRAM bank in a single cycle.  The on-FPGA SRAM banks are dual-port: reads are from port A; writes are to port B.  The primary computing units are shown in blue -- each is a \textit{software} thread in the software implementation.  Observe that there are 4 instances of 5 different compute units: 20 units (threads) in total.  Edges in the figure represent multiple FIFO queues for communication of data/control between the units; queues are not shown for clarity.

The high-level behavior is as follows: The \textit{data-staging/control} units receive an instruction from the ARM processor to perform convolution, padding, or max-pooling. We do not focus on fully connected layers, since it is essentially matrix multiplication and most CNN computational work comprises convolution. Although the padding and subsampling operations can be performed by a processor fairly efficiently, in most CNNs, they are tightly interleaved with convolution operations. Supporting them in hardware minimizes memory traffic between the FPGA and HPS.

For a convolution instruction, four tiles from different output feature maps (OFMs) are computed simultaneously.  The four concurrently computed OFM tiles are at the same $x/y$ location. Each data-staging/control unit loads a subset of input feature maps (IFMs) from an on-FPGA SRAM bank, as well as corresponding filter weight data from four filters.  On each clock cycle weights and IFM data are injected into the \textit{convolution} units.  Each convolution unit performs 64 multiply operations each clock cycle, thus the entire accelerator performs 256 multiplication operations per cycle.  Products from the convolution units are sent to the \textit{accumulator} units (center of the figure).  Each accumulator unit is responsible for maintaining the values of one tile (16 values) in an OFM. In the figure, for clarity, some edges between convolution units and accumulator units are omitted.  When an OFM tile is completed, it is sent to the \textit{write-to-memory} unit and written to an on-FPGA SRAM bank.

For a padding or max-pooling instruction, the data-staging/control units send IFM data and instructions to the \textit{pool/pad} units, capable of performing \textit{any} style of padding/max-pooling, described below.  Pooled/padded tiles are then forwarded to the write-to-memory units and written to SRAM banks.  Padding/pooling of four OFM tiles is done concurrently.

\subsection{Convolution and Zero-Weight Skipping}

OFMs are computed on a tile-by-tile basis to completion without any intermediate swap-out to off-chip memory in an output-stationary manner. This style allows us to keep a fixed datapath width and not compromise accuracy by rounding partial sums. The convolution unit contains a computational sub-module that multiplies one weight per clock cycle to 16 IFM values and accumulates the resulting 16 products to the corresponding 16 OFM values in an OFM tile being computed.  

\begin{figure}
  \centering
    \includegraphics[width=0.7\linewidth]{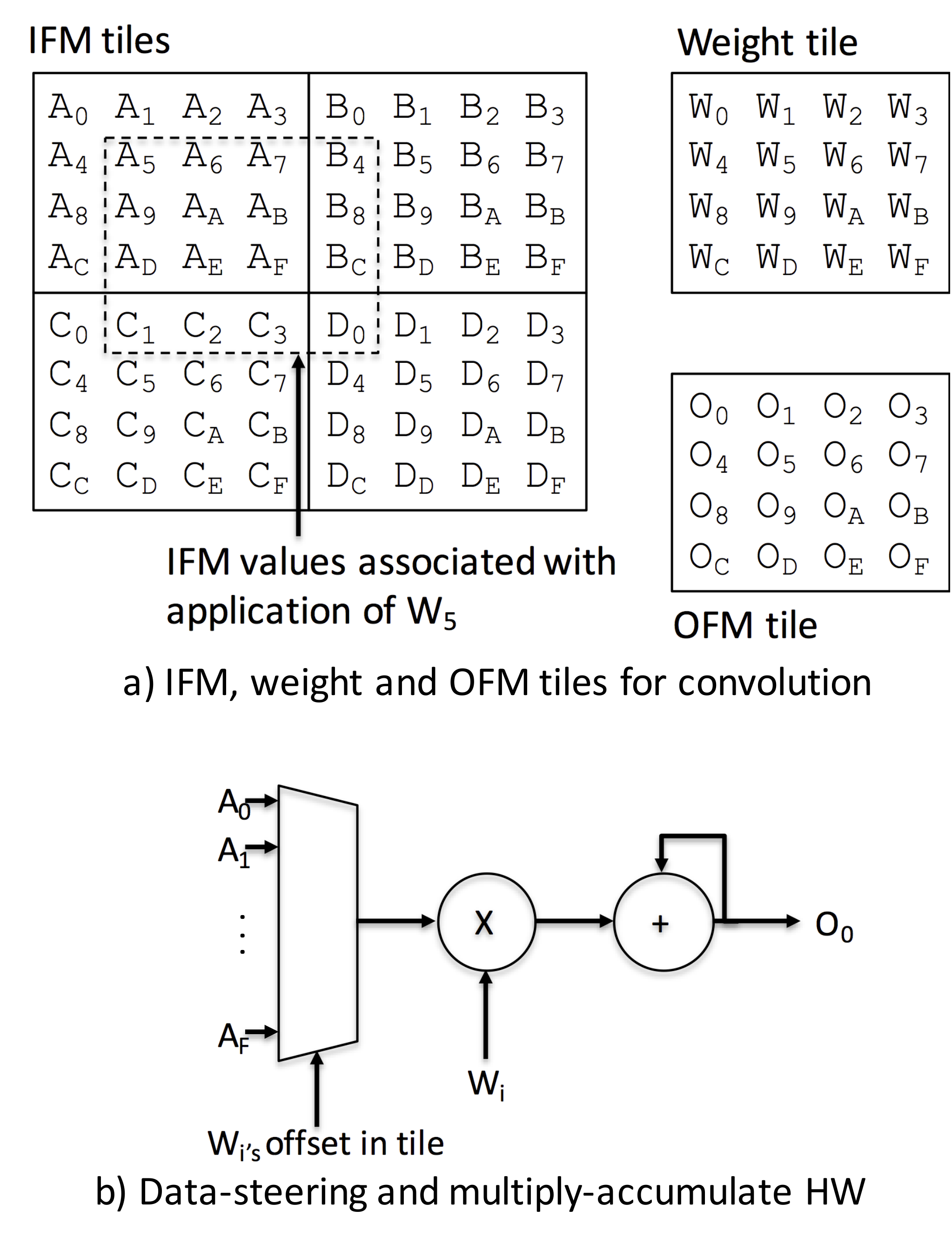}
    \caption{Data and architecture for convolution and zero-weight skipping.}
    \label{fig:zero}
   \vspace{-0.2in}
\end{figure}

Fig.~\ref{fig:zero}(a) shows an OFM tile (lower right), a weight tile (upper right), and four contiguous IFM tiles.  For this example, assume that the upper-left IFM tile is $x/y$ aligned with the OFM tile being computed and that the filter dimensions are smaller than $4 \times 4$ (the tile size).
In a given clock cycle, a weight in the weight tile is selected and multiplied by 16 IFM values.  The example illustrates, with a dotted rectangle, the region of 16 IFM values with which the weight $W_5$ is multiplied.  Observe that the intra-tile $x/y$ offset of $W_5$ defines the region of IFM values which with it is multiplied.  This produces 16 products: $W_5 \cdot A_5$, $W_5 \cdot A_6$ $\dots$ $W_5 \cdot D_0$.  The products are accumulated to the corresponding OFM values: $O_0$, $O_1$, $\dots$, $O_F$, respectively.  

Fig.~\ref{fig:zero}(b) shows steering and multiply-accumulate hardware for value $O_0$ -- the top-left value in an OFM tile being computed.  Registers are omitted for clarity. When a weight $W_i$ is selected from the weight tile, the specific IFM value with which the weight $W_i$ is multiplied depends on $W_i$'s offset within the weight tile.  For the OFM value $O_0$, this may be any of the IFM values $A_0$, $A_1$, $\dots$, $A_F$, shown on the data inputs of the multiplexer, while the select inputs receive the weight's offset within the weight tile.   

With the proposed tiling and convolution approach, zero-weight skipping is straightforward.  For a given neural network model, the non-zero weights and their intra-tile offsets are ``packed'' offline in advance in software.  The packing procedure only needs to be done once for a given CNN model such as VGG-16.  During inference, the accelerator receives 
the weight values and their intra-tile offsets in a packed format that is read directly into scratchpad memory. One non-zero weight is applied per clock cycle; no cycles are spent on weights having a value of 0.  

\subsubsection{Scaling Out Convolution}

Each of the four data-staging/control units in Fig.~\ref{fig:block_dia} manages one quarter of the IFMs and corresponding weights.  Every clock cycle, a data-staging/control unit injects IFM data and four weights (from four different filters) and their respective weight offsets into a convolution unit.  Each of the four weights is multiplied by 16 IFM values, as described above.  Thus, each convolution unit performs $4 \times 16 = 64$ multiplications/cycle.  
While the weights from a tile are being applied to (i.e.~multiplied with) IFM data, the \textit{next} required tiles of IFM data are simultaneously preloaded from the on-FPGA SRAM banks.  Fig.~\ref{fig:zero} shows that four IFM tiles are needed to apply a weight tile and hence, since one tile/cycle can be loaded from an SRAM bank, at \textit{least} four clock cycles must be spent processing a weight tile.
This restriction implies that the upper-bound cycle-count reduction from zero-skipping is $(16-4)/16$ = 75\% in our implementation.
Moreover, note that OFMs being computed simultaneously may have different numbers of non-zero weights in their filters, causing pipeline bubbles and reduced efficiency. The completion of all four OFM tiles at a given $x/y$ tile position is synchronized using a Pthreads barrier. 

\subsection{Padding and Pooling}

\begin{figure}
  \centering
    \includegraphics[width=0.8\linewidth]{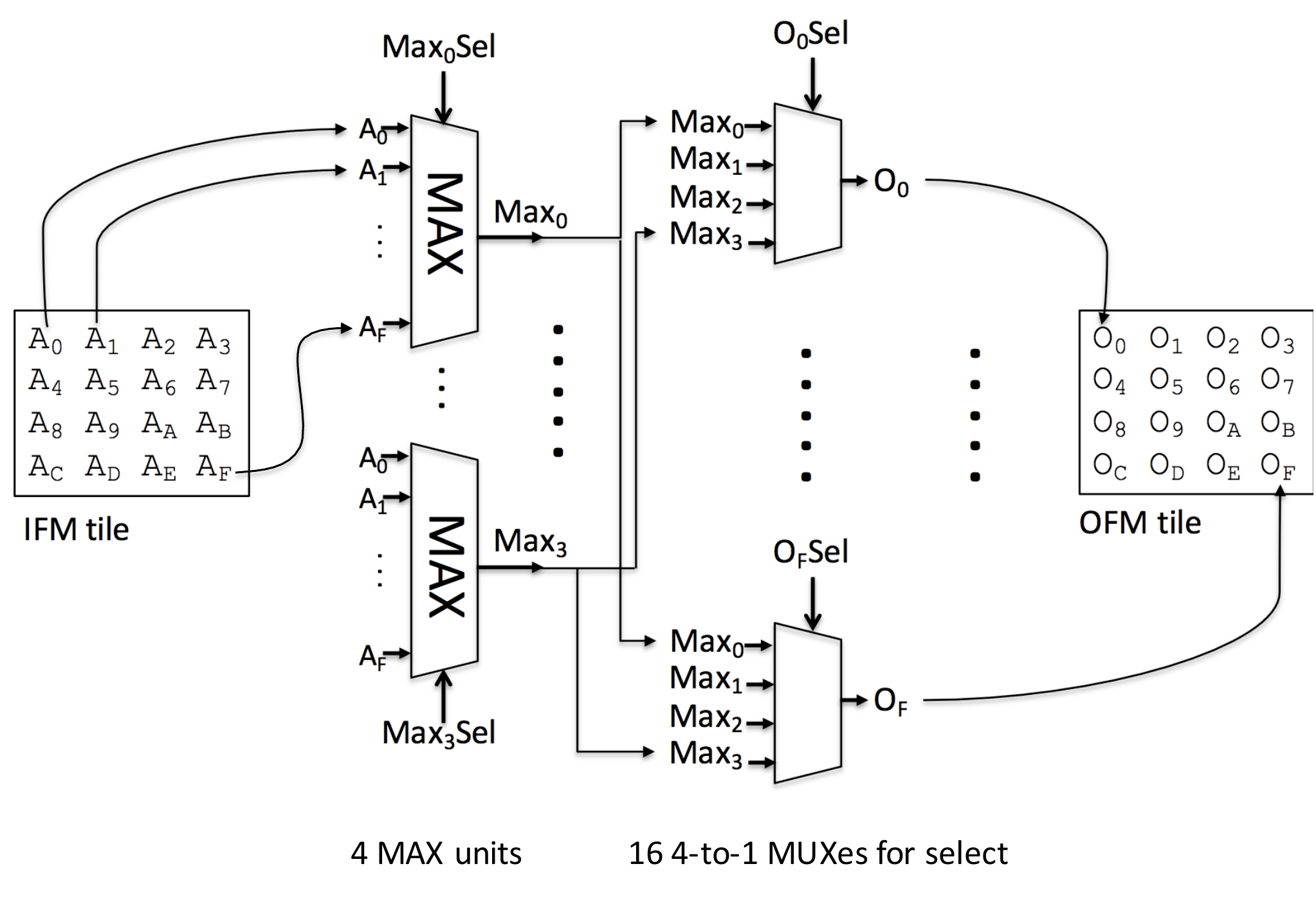}
    \caption{Hardware architecture of padding/pooling unit.}
    \label{fig:padpool}
    \vspace{-0.2in}
\end{figure}

Fig.~\ref{fig:padpool} shows the padding/max-pooling unit.  The controller/data-staging unit injects an IFM tile (shown on the left), as well as an \textit{instruction} that specifies the desired padding/max-pooling behavior.  There are four MAX units that, based on the instruction, select the maximum of \textit{any} of the 16 IFM values in the input tile.   The MAX units feed 16 multiplexers: one for each value of the OFM tile being computed.  Based on the instruction, each value in the OFM tile may be updated with one of the MAX unit outputs, or alternately, may retain its old value.  

To implement padding, which does not involve taking the maximum of multiple IFM values, %
the MAX units return a \textit{single} value from the IFM (i.e.~find the maximum among a single value).   The specific number of MAX units (four in this case), is inspired by the needs of VGG-16, which requires $2 \times 2$ max-pooling regions with a stride of 2.  However, with just a few instructions, the padding/max-pooling unit is capable of realizing \textit{any} padding/max-pooling layer (e.g.~a variety of max-pooling region sizes or strides).   Moreover, since all units are described in software, it is straightforward to increase the number of MAX functional units within the padding/max-pooling unit.

\section{Implementation}

\subsection{High-Level Synthesis}

Use of the LegUp HLS Pthreads flow permitted accelerator design, development and debugging to proceed in software. Debug and test were simplified, as the parallel software execution aligns closely with the target hardware architecture.

The primary HLS constraints applied were loop pipelining, if-conversion, automated bitwidth minimization~\cite{Gortbit}, and clock-period constraints. To achieve optimized loop pipelines (initiation interval [II] = 1), it was necessary to remove control flow from the \texttt{C} code to the extent possible by making use of the ternary operator (\texttt{<cond> ? <val1> : <val2>}) to implement MUX'ing instead of using conditionals.  The \texttt{C} specification of the accelerator is $\sim$5,600 LOC.  The Verilog RTL automatically produced by LegUp was modified (using a script) in three ways: 1) pragmas were added to direct the Intel synthesis tools to implement the FIFO queues using LUT RAM instead of block RAM (saving precious block-RAM resources); 2) the on-FPGA SRAM banks were brought to the top-level of the hierarchy, making them accessible to the DMA unit for writing/reading; and 3) the port assignment for the on-FPGA RAM banks was altered so that reads and writes have exclusive ports, reducing contention/arbitration. Our design goal was skewed towards maximizing performance; however, it is possible to apply directives in LegUp that lower resource utilization at the expense of some performance. 

A challenge with HLS is the disconnect between the software source code and the generated hardware -- it can be difficult for one to know how to change the source code to effect desired changes in the hardware.  In this work, this issue arose in the context of the controller/data-staging unit, which synthesized to a module with a large number of FSM states (hundreds), and consequent high-fanout signals (e.g.~the FSM stall logic).  To mitigate this, we split the unit into two \texttt{C} functions, one for convolution instructions and one for padding/max-pooling instructions, each having a simpler FSM than the original monolithic controller.

With reference to Fig.~\ref{fig:block_dia}, the pool/pad, convolution, write-to-memory, and accumulator units are all streaming kernels generated by LegUp HLS that can accept new inputs every clock cycle (II = 1).  The data-staging/control unit is  capable 
of injecting data into the compute units every cycle.  We reinforce that the \textit{entire} accelerator, including the compute pipelines and the relatively complex control as described in Section~\ref{sec:arch}, is \textit{synthesized} to Verilog RTL from parallel \texttt{C} software.  Manual RTL design was used \textit{solely} for the DMA unit. While streaming audio/video applications are typically emblematic as being ``ideal'' for the use of HLS, here, we show that HLS can effectively be used to generate sophisticated control and data-staging for streaming hardware.

\subsection{VGG-16 Reduced Precision and Pruning}

Beginning with the pre-trained VGG-16 model~\cite{DBLP:journals/corr/SimonyanZ14a}, we increased the sparsity by pruning and reduced the precision to 8-bit magnitude-plus-sign representation by scaling. Pruning and precision reduction were done using Caffe, in a manner similar to~\cite{DBLP:journals/corr/HanMD15}.  We consider two VGG-16 models: 1) with reduced precision and 2) with reduced precision and pruning.  With variant \#2, inference accuracy in validation was within 2\% of the original unpruned floating point, which can be improved further through training. %
\subsection{Software}

Software executing on the on-chip ARM processor handles the loading and pre-processing of network weights, biases and test images. Pre-processing includes the reordering of data into tiled format for our accelerator. The framework sends the instruction and calls the hardware driver for inference.

\subsection{System Integration, Accelerator Scale-Out}

The accelerator core, DMA controller, and host processor communicate via an interconnect network synthesized using Intel's \textit{Qsys} System Integration tool. Two separate systems are instantiated: \emph{System I} is a high bandwidth 256-bit bus that performs DMA to and from system DRAM to the accelerator banks. \emph{System II} is a set of Avalon Memory-Mapped (AMM) interfaces between the host ARM processor and control and status registers on the accelerator core and DMA unit.  The accelerator and DMA unit are controlled using \emph{System II} by the host ARM processor.

We target a mid-range-sized Intel Arria 10 SX660 FPGA, whose size permits us to instantiate two instances of the accelerator shown in Fig.~\ref{fig:block_dia}, where each instance operates concurrently on separate stripes of FMs (see Fig.~\ref{fig:tile}).  The overall multi-accelerator system is capable of 512 MACs/cycle.    

\section{Experimental Study}

A unique advantage of HLS is that one can synthesize multiple architecture variants from software and constraint changes alone.  We analyze area, power and performance 
for four architecture variants running the two VGG-16 CNN models mentioned above: reduced precision without and with pruning (higher fraction of zero weights).  
The four architecture variants considered are as follows (labels  in brackets): 

\begin{enumerate}
\item A non-optimized simplified accelerator variant with a single convolution sub-module capable of performing at most 16 MACs/cycle  (16-unopt).  
\item A non-performance-optimized variant with one instance of the accelerator in Fig.~\ref{fig:block_dia} capable of performing at most 256 MACs/cycle (256-unopt).  
\item A performance-optimized variant of Fig.~\ref{fig:block_dia} (256-opt).
\item A variant with two instances of the accelerator in Fig.~\ref{fig:block_dia} capable of performing at most 512 MACs/cycle (512-opt).
\end{enumerate}

The 16-unopt architecture computes a single OFM tile at a time, and consequently requires no synchronization among multiple control/data-staging units. Analysis of the 16-unopt architectures gives insight into the HLS hardware quality in the absence of synchronization overhead. Both the 16-unopt and the 256-unopt architectures were not performance-optimized and as such, they consume minimal area and, to verify functional correctness, were clocked at 55MHz.  To produce higher-performance variants, we tightened the clock-period constraint supplied to the LegUp HLS tool, and also invoked performance optimizations in the Intel RTL-synthesis tool: retiming, physical synthesis, higher place/route effort.  The 256-opt and 512-opt architectures were clocked at 150 MHz and 120 MHz, respectively.  Routing of the 512-opt architecture failed at higher performance targets due to high congestion.

\begin{figure}
  \centering
    \includegraphics[width=0.8\linewidth]{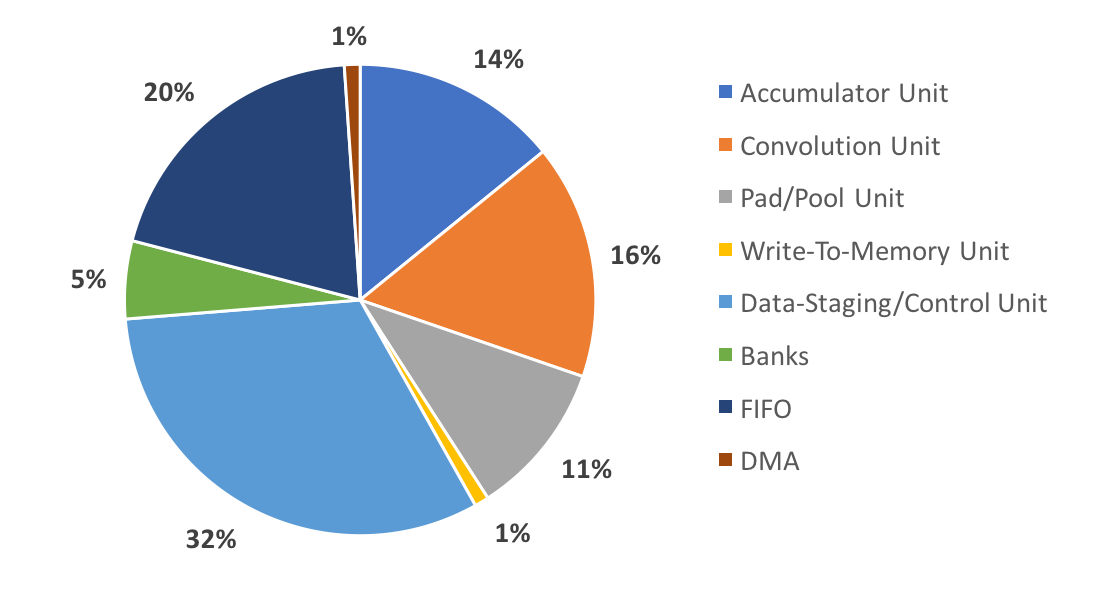}
    \caption{ALM usage by each unit in the accelerator.}
    \label{fig:alm_area_breakdown}
    \vspace{-0.2in}
\end{figure}

Intel FPGAs have 3 main types of resources: Adaptive Logic Modules (ALMs -- lookup-table-based logic), DSP and RAM blocks. Our 256-opt accelerator uses 44\% of the ALM logic, 25\% of the DSP and 49\% of the RAM blocks. %
Fig.~\ref{fig:alm_area_breakdown} shows the breakdown of ALM usage for each module. The convolution, accumulator and data-staging/control modules take up most of the area, due to the heavy MUX'ing required in these units. Most of the DSP blocks are used in the convolution and accumulator modules.
We adjust the RAM block usage to maximize our bank size given the number of available RAMs. 

\begin{figure}
  \centering
    \includegraphics[width=1\linewidth]{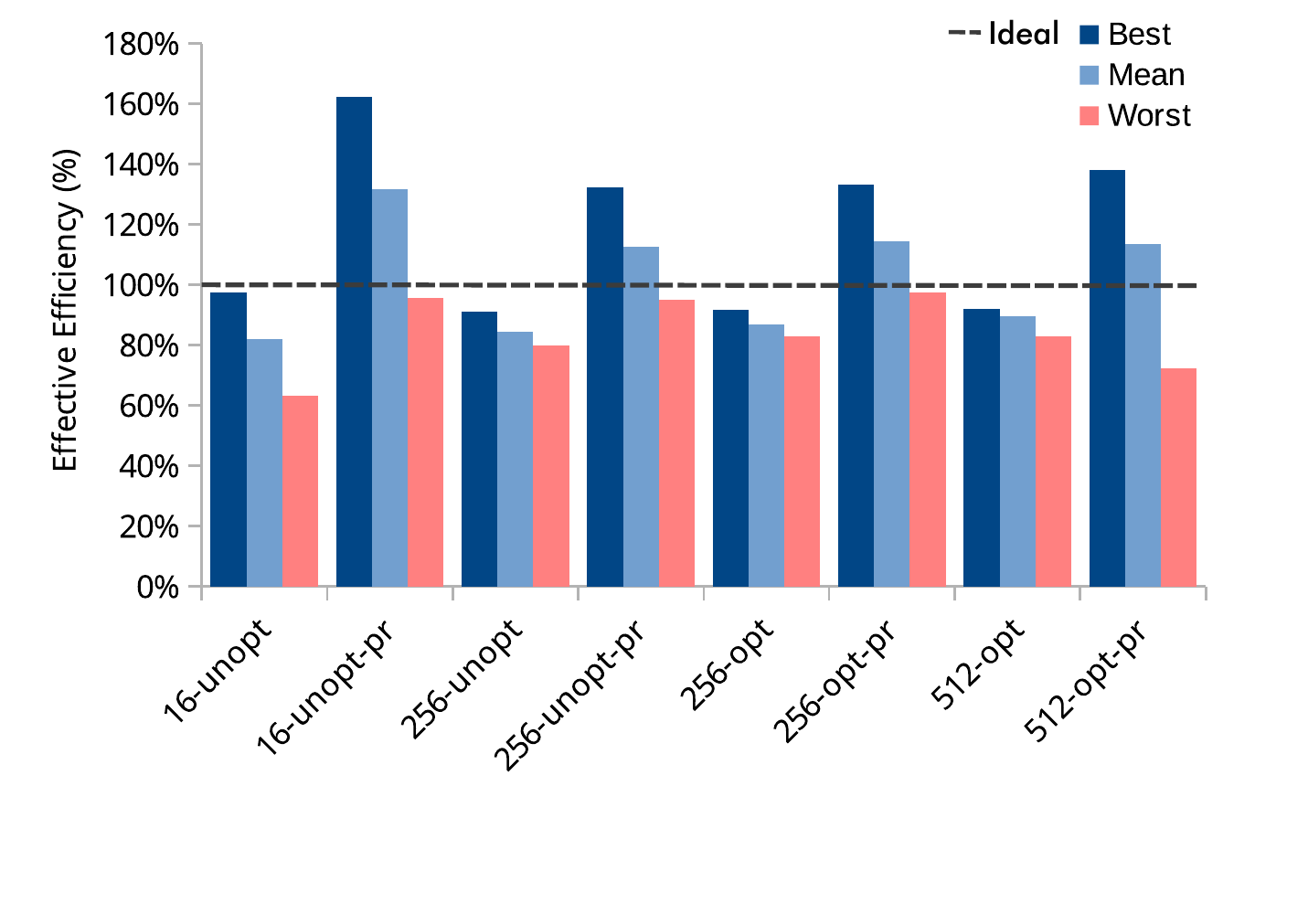}
    \vspace{-0.55in}
    \caption{Efficiency of each accelerator variant for VGG-16 inference.}
    \label{fig:eff}
\end{figure}

We consider the efficiency of the HLS-generated hardware by comparing the experimentally observed throughput ($ops/elapsed\,  time$) with the theoretically minimum \textit{ideal} throughput numbers. Ideal throughput is defined as \textit{peak throughput * total number of computations}. We add an overhead ($\sim$15\% but varies by layer) for the increased number of MAC operation due to limited on-FPGA SRAM bank size -- ``striping''.  Results in Fig.~\ref{fig:eff} illustrate the efficiency of various architectures with the pruned and unpruned VGG-16 model. Results obtained using a pruned network are labeled as ``-pr''. ``Best'' and ``worst''  refer to the highest and lowest throughput for any single convolutional layer of VGG-16, respectively. Mean refers to the average throughput across all VGG-16 layers. The \textit{ideal} throughput value is indicated as a dotted line on Fig.~\ref{fig:eff}.

The underlying reason for differences between best, worst, and mean is that, for deeper layers of VGG-16, the ratio of weight data to FM data increases, imposing a higher overhead for unpacking weights and offsets in our accelerator, reducing effective throughput. Using the pruned network we see greater than 100\% efficiency, due to the zero-skipping avoiding some multiply-accumulates altogether. For the non-pruned VGG-16, we are not far from the ideal throughput -- usually within $\sim$10\%.  
This analysis shows the HLS-generated hardware is quite efficient; overhead from HLS is minimal from the cycle latency perspective. 

\begin{figure}
  \centering
    \includegraphics[width=1\linewidth]{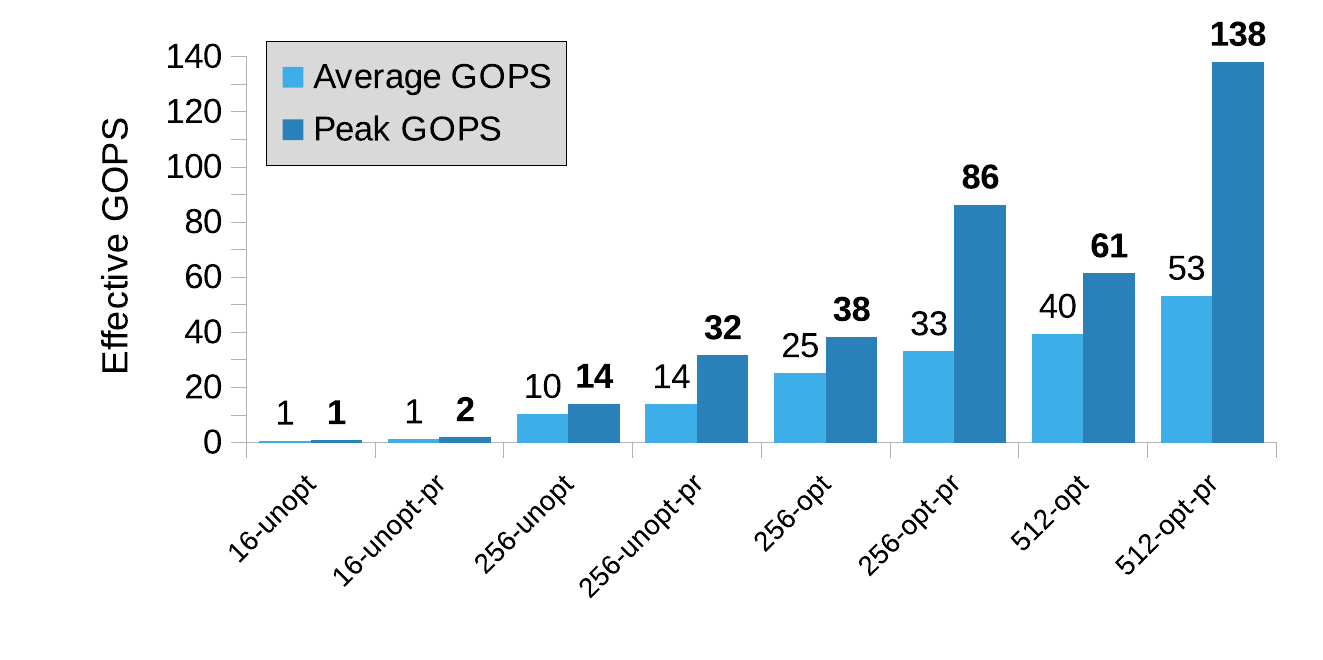}
    \vspace{-0.2in}
    \caption{Absolute GOPS/s across accelerator variants for VGG-16.}
    \label{fig:abs_ops}
    \vspace{-0.2in}
\end{figure}
Secondly, we look at our results in terms of absolute performance (GOPS). Fig.~\ref{fig:abs_ops} shows that our 512-opt accelerator achieved the highest average (peak) throughput of 39.5 GOPS (61 GOPS), and with the pruned network, the effective average (peak) performance increases to 53.3 GOPS (138 GOPS). We can clearly see the effects of zero-skipping in these results. By pruning the network, we were able to increase our performance by $\sim$1.3$\times$, on average, and $\sim$2.2$\times$ in the peak case. %
The peak performance numbers are considerably higher than the average in the pruned case, as peak throughput requires uniformly sparse filters applied concurrently for even workload balancing. Future work could include grouping filters in advance according to similarity in non-zero-entry counts to maximize available zero skipping and balance the work.

While the absolute performance numbers are relatively modest and in line with prior work (cf.~Section~\ref{sec:related}), the results in Fig.~\ref{fig:abs_ops} underscore a key advantage of HLS, namely, a wide range of architectures with distinct performance/area trade-offs can be produced by software and HLS constraint changes alone. For example, in the \textit{opt} vs.~\textit{unopt} variants, the clock-period constraint applied in HLS impacts the degree of pipelining in the compute units and control. It would be expensive and time-consuming to produce hand-written RTL for all  architecture variants considered.  
Finally, we note that on a larger Arria 10 FPGA family member (e.g.~GT1150), with nearly double the capacity, software changes alone would allow us to scale out the design further.

Power consumption measurements are given in Table \ref{tbl:power}. All measurements are peak power measured while running the accelerator on the worst-case VGG-16 layer. A \emph{board}-level power measurement is provided, alongside the power consumption of the FPGA by itself.  Power numbers in parentheses are dynamic power; numbers outside parentheses are static + dynamic power.

\begin{table}
    \centering
    \caption{Power Consumption}
    \label{tbl:power}                \resizebox{\columnwidth}{!}{%
        \tiny
        \begin{tabular}{@{}cccll@{}}
        Accelerator Variant & Peak Power (mW) & GOPS/W & GOPS/W (peak) \\ \midrule
        256-opt (FPGA)  & 2300 (500)  & 13.4 & 37.4\\ 
        512-opt (FPGA)  & 3300 (800)  & 13.9 & 41.8\\
        256-opt (Board) & 9500        & 3.5  & 9.05\\ 
        512-opt (Board) & 10800       & 5.6  & 12.7\\ 
        \end{tabular}
    }
    \begin{flushleft}
        \textit{\small*dynamic power is parenthesized} 
    \end{flushleft}
    \vspace{-0.25in}
\end{table}
\section{Related Work}\label{sec:related}

Recent years have seen considerable research on custom hardware accelerators for deep CNN inference designed in manual RTL, realized in ASICs  (e.g.~\cite{DBLP:journals/jssc/ChenKES17,DianNao2014}) and FPGAs (e.g.~\cite{Eriko17}).   The works most comparable to ours use HLS and target FPGAs.  \cite{Cong15} applied Xilinx's Vivado HLS to synthesize an accelerator for the convolution operation, achieving 61~GFLOPS peak performance. Suda et~al.~\cite{Suda16} synthesized a CNN accelerator from OpenCL, achieving peak performance of 136~GOPS for convolution.  In~\cite{2017arXiv170103534A}, Intel synthesized a CNN accelerator from OpenCL that incorporates the Winograd transform, half-precision floating point, and achieves over 1~TFLOPS performance. In~\cite{2016arXiv161207119U}, Xilinx synthesized significant parts of a binarized CNN accelerator with Vivado HLS.  In binarized CNNs, both weights and activations are represented by 1-bit values.  OpenCL along with Xilinx's SDAccel OpenCL HLS tool was also used in~\cite{DBLP:journals/corr/DiCeccoLVCTA16} to synthesize a single-precision floating-point CNN accelerator incorporating Winograd and achieving 50~GFLOPS performance.  The latter work also provides integration with the Caffe framework.  

Generally, the OpenCL implementations above use PCIe connectivity with a host processor and are more suited for data center applications; whereas, our system is intended for embedded. To the best of the authors' knowledge, our work is the first to synthesize a CNN accelerator from parallelized \texttt{C}-language software that incorporates a novel zero-skipping approach and reduced precision, and illustrates how, beginning with a software-parallelized \texttt{C} specification of the architecture, constraints to the HLS and RTL synthesis tools can be applied to generate a variety of accelerator architectures with different performance/area trade-offs.

\section{Conclusions and Future Work}

An FPGA-based CNN inference accelerator was synthesized from parallel \texttt{C} software using the LegUp HLS tool, incorporating a zero-weight skipping architecture and reduced-precision arithmetic. The end-to-end system contains a dual-core ARM processor, the accelerator, on-FPGA memories (backed by DDR4) and DMA.  Software running on the ARM issues instructions to the accelerator for convolution, padding and pooling.  Complex datapath and control logic was synthesized entirely from \texttt{C}, and the use of HLS permitted a range of architectures to be evaluated, through software and constraint changes alone.  Future work involves the use of HLS to synthesize accelerators for other neural network styles, including binarized, ternary and recurrent networks.

\def\bibfont{\scriptsize}
\bibliography{articles.bib}
\bibliographystyle{IEEEtran}
\bibliographystyle{abbrv}
\bibliographystyle{unsrt}
\end{document}